\documentclass{article}

\usepackage[square,sort,comma,numbers]{natbib}

\usepackage{amsmath}
\usepackage{graphicx}
\usepackage{arxiv}

\usepackage{setspace}

\usepackage[utf8]{inputenc} % allow utf-8 input
\usepackage[T1]{fontenc}    % use 8-bit T1 fonts
\usepackage{hyperref}       % hyperlinks
\usepackage{url}            % simple URL typesetting
\usepackage{booktabs}       % professional-quality tables
\usepackage{amsfonts}       % blackboard math symbols
\usepackage{nicefrac}       % compact symbols for 1/2, etc.
\usepackage{microtype}      % microtypography
\usepackage{lipsum}
\usepackage{mathtools}
\newcommand{\norm}[1]{\left\lVert#1\right\rVert}

\title{A Novel Method of Extracting Topological Features from Word Embeddings}
\author{
  Shafie Gholizadeh \\
  Department of Computer Science \\
  University of North Carolina at Charlotte\\
  Charlotte, NC 28223 \\
  \texttt{sgholiza@uncc.edu} \\
   \And
  Armin Seyeditabari \\
  Department of Computer Science\\
  University of North Carolina at Charlotte\\
  Charlotte, NC 28223 \\
  \texttt{sseyedi1@uncc.edu@uncc.edu} \\
   \And
 Wlodek Zadrozny \\
  Department of Computer Science\\
  University of North Carolina at Charlotte\\
  Charlotte, NC 28223 \\
  \texttt{wzadrozn@uncc.edu} \\
}

\begin{document}
\maketitle

\onehalfspacing

\begin{abstract}
In recent years, topological data analysis has been utilized for a wide range of problems to deal with high dimensional noisy data. While text representations are often high dimensional and noisy, there are only a few work on the application of topological data analysis in natural language processing. In this paper, we introduce a novel algorithm to extract topological features from word embedding representation of text that can be used for text classification. Working on word embeddings, topological data analysis can interpret the embedding high-dimensional space and discover the relations among different embedding dimensions. We will use persistent homology, the most commonly tool from topological data analysis, for our experiment. Examining our topological algorithm on long textual documents, we will show our defined topological features may outperform conventional text mining features.
\end{abstract}

% keywords can be removed

Topological Data Analysis (TDA) is the collection of mathematical tools to define and capture shapes, and then analyze the structure of shapes in the data. Recently TDA has been applied on variety of problems dealing with numeric data, specially when a noisy data set can be represented by a high dimensional data cloud. But to deal with textual data, it would be a complicated problem to define and analyze topological space over a corpus. In this case we need to choose a numerical latent representation of the text and then measure its topological properties.

In this paper we will introduce a novel method to extract topological features from word embedding representations of textual document utilizing \emph{persistent homology} and show how to use those topological features for text classification. We will discuss under which circumstances, extracting topological features might be useful for text classification. Especially, for long textual documents, our defined topological features can outperform conventional text mining features. For this study, we will use persistent homology, the most commonly tool from topological data analysis.

The rest of the paper is organized as follows: In Sect. \ref{sec:background}, we will introduce fundamental definitions in TDA and some contributions of TDA in machine learning and natural language processing. Sect. \ref{sec:method} contains the details of our topological method for text representation. The descriptions of the data sets on which we examine our topological method are mentioned in Sect. \ref{sec:data} followed by experimental results and discussion in Sect. \ref{sec:results}.

\section{Background}
\label{sec:background}

\begin{figure*}[!ht]
\centering
\includegraphics[width = 0.99\textwidth]{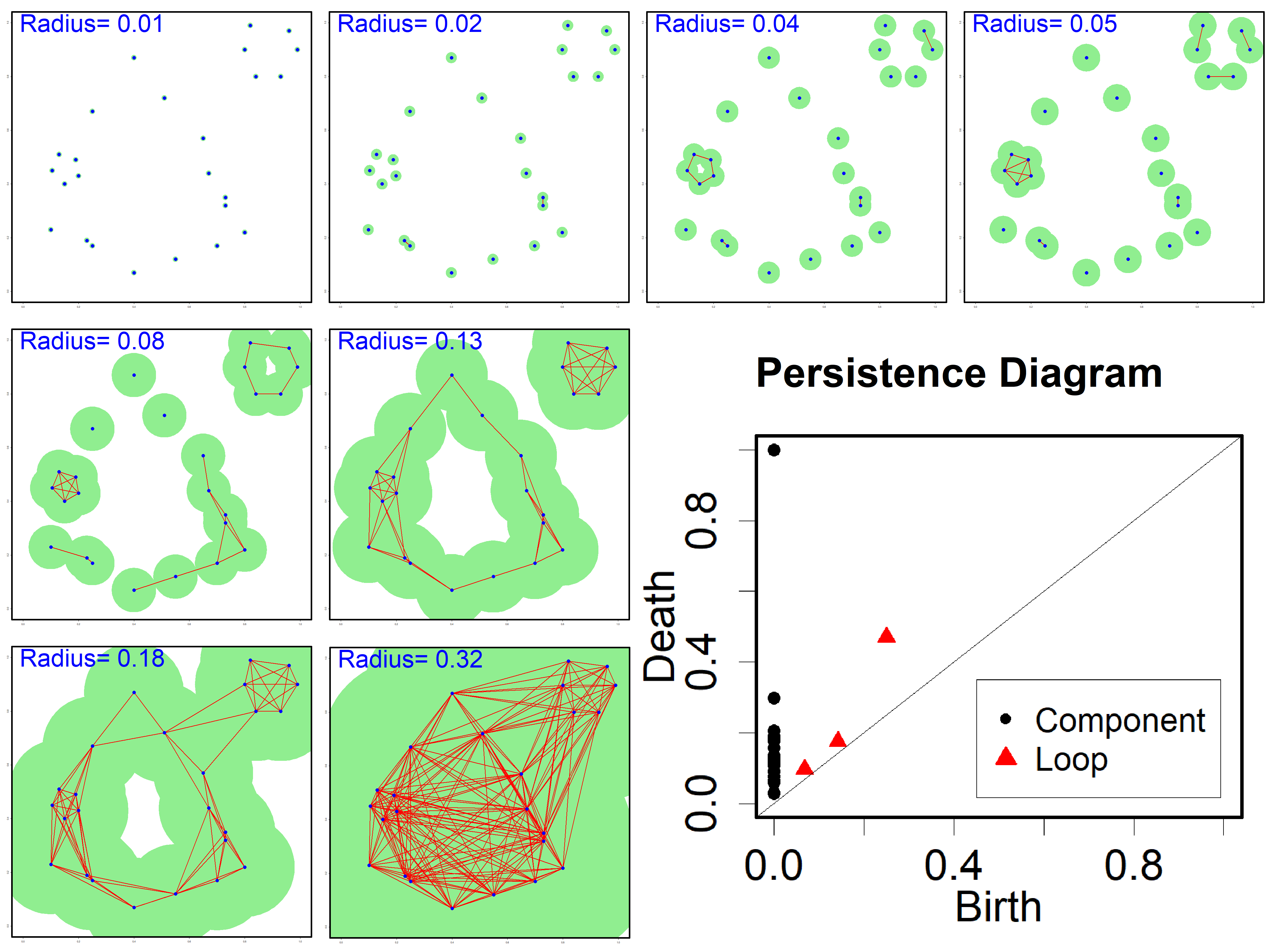}
\caption{Persistent Homology}
\label{fig:ph}
\end{figure*}

Persistent homology \cite{edelsbrunner2000topological,carlsson2009topology, edelsbrunner2008persistent, chen2015mathematical} is a technique in TDA for multi-scale analysis of data clouds.  In a data cloud, there are no pre-defined links between the points. Therefore, there are no $k$-simplices except $0$-simplices each of which trivially referring to a single data point. Betti numbers are all zero, except $\beta_0$ which is equal to the number of data points. The main idea of persistent homology is to define an $\epsilon$-distance around each data points and then connect those points with overlapping $\epsilon$-distances. We may change $\epsilon$ in the range of $(0, +\infty)$. Setting $\epsilon = 0$, there will be no link between the points. However, increasing $\epsilon$ gradually, some points will get connected. When $\epsilon$ is large enough, ($\epsilon \rightarrow \infty$), all data points are connected to each other--- i.e. $n$ data points will consist an $n$-simplex. In the way to increase $\epsilon$ from $0$ to $+\infty$, number of connected components will change. Also, many loops in the data may appear and disappear---i.e., Betti numbers are changing. On a data cloud, changes in Betti numbers are captured by persistent homology. More precisely, Persistence Diagram \cite{edelsbrunner2000topological} captures the birth and death $\epsilon$'s of each component, loop, void, etc. An example for persistent homology on a data cloud along with the resulted persistence diagram are illustrated in Fig. \ref{fig:ph}. Alternatively, we may show the birth and the death of loops with barcodes where each hole is represented by a bar from its birth $\epsilon$ to its death $\epsilon$ \cite{collins2004barcode,carlsson2005persistence, ghrist2008barcodes}. For a more general review of persistent homology and its applications we refer the reader to \cite{edelsbrunner2008persistent,zomorodian2005computing,munch2017user, gholizadeh2018short}.

\begin{figure}[!ht]
\centering
\includegraphics[width = 0.45\textwidth]{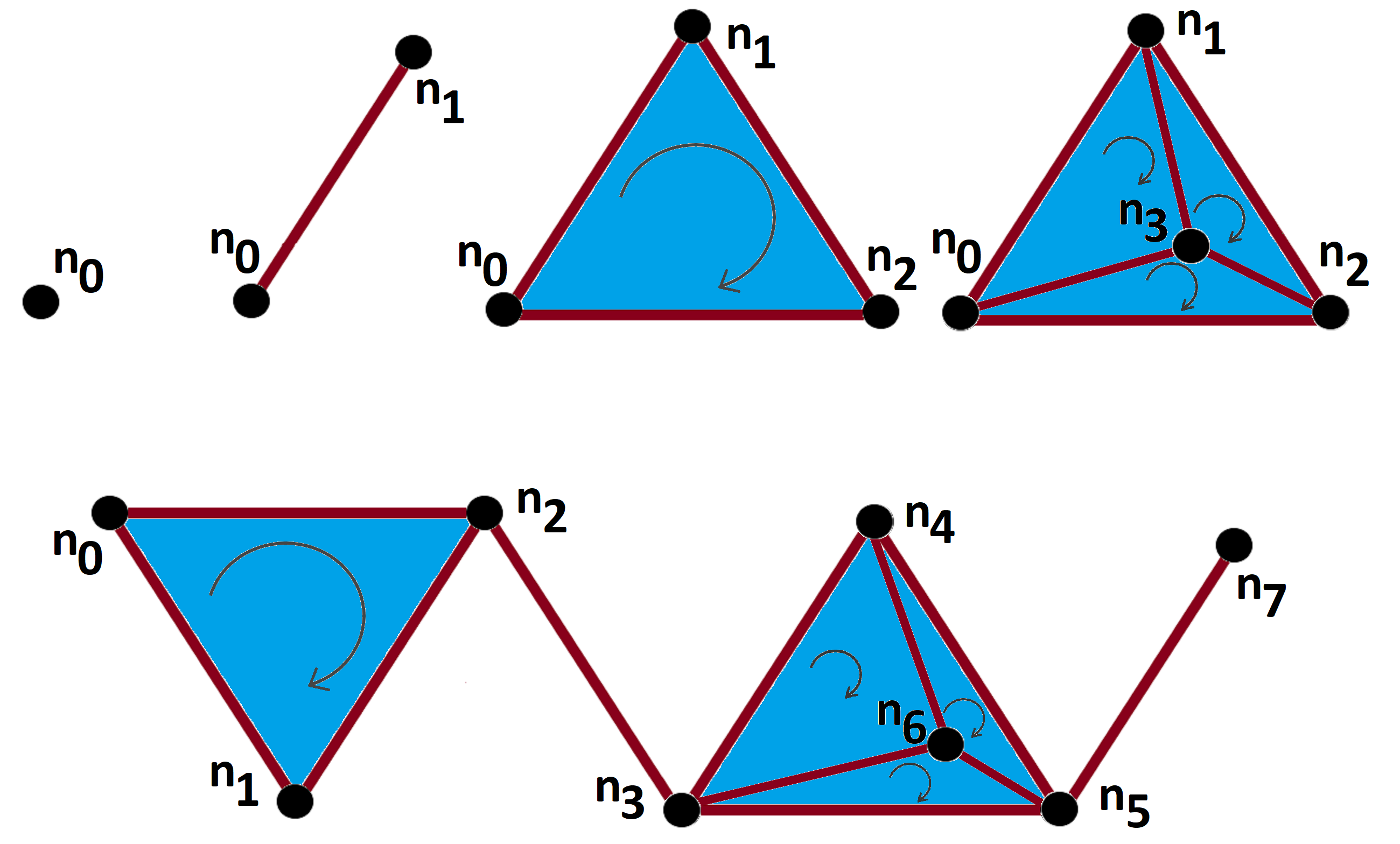}
\caption{$0$-simplex, $1$-simplex , $2$-simplex and $3$-simplex (top). An example of simplicial complex (bottom).}
\label{fig:SimpComplex}
\end{figure}

While TDA and specially persistent homology are applied on a wide range of problems (e.g., system analysis \cite{maletic2016persistent, garland2016exploring, pereira2015persistent, khasawneh2014stability, perea2015sliding, stolz2017persistent}, network coverage \cite{de2006coordinate, de2007coverage}, etc),  there are only a few studies using them for natural language processing. Zhu in \cite{zhu2013persistent} used persistent homology to find repetitive patterns in the text, comparing vector space representations of different blocks in the documents. Doshi and Zadrozny in \cite{doshi2018movie} utilized Zhu's algorithm \cite{zhu2013persistent} for movie genre detection on the IMDB data set of movie plot summaries. The authors showed how persistent homology can improve the classification. In \cite{wagner2012computational}, Wagner et al. utilized TDA to measure the discrepancy among documents represented by their TF-IDF matrices. Guan et al. in \cite{guan2016topological} utilized topological collapsing algorithm \cite{wilkerson2013simplifying} to develop an unsupervised method of key-phrase extraction. Almgren et al. in \cite{almgren2017mining} and \cite{almgren2017extracting} examined the feasibility of persistent homology for social network analysis. The authors utilized Mapper algorithm \cite{singh2007topological} to predict image popularity based on word embedding representations of images' captions. In \cite{torres2015topic}, Torres-Tram{\'o}n et al. introduced a topic detection algorithm for Twitter data analysis  utilizing Mapper algorithm to map the term frequency matrix to the topological representations. Savle and Zardozny in \cite{savle2019topological} used TDA to study discourse structures and showed that topological features derived from the relation of sentences can be informative for prediction of text entailment. Considering the word embedding representation of the text as high-dimensional time-series, some ideas from the recent applications of persistent homology in time series analysis \cite{pereira2015persistent, khasawneh2014stability, perea2015sliding, maletic2016persistent, stolz2017persistent} are also considerable for text processing.

\begin{figure}[!ht]
\centering
\includegraphics[width = 0.47\textwidth]{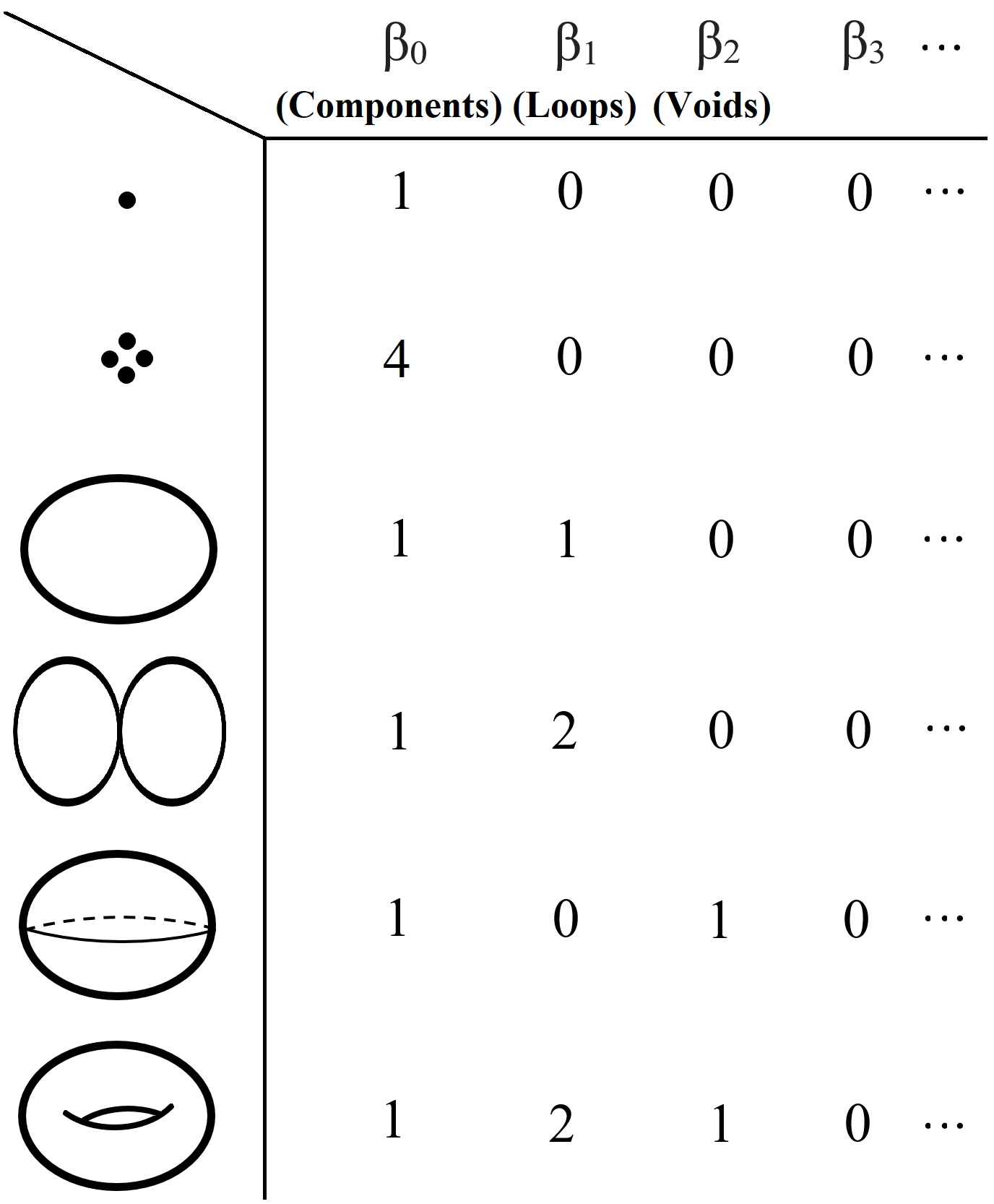}
\caption{Betti numbers for some simple shapes.}
\label{fig:Betti}
\end{figure}

In our prior work \cite{gholizadeh2018topological}, we applied persistent homology on a set of novels and tried to predict the author without using any conventional text mining features. For each novel, we built an adjacency matrix whose elements are measuring the co-appearance of the main characters (persons) in the novel. Utilizing persistent homology, we analyzed on the graph of the main characters in the novels and fed the resulted topological features (conveying topological styles of novelists) to a classifier. Despite the novelty of the algorithm (using persistent homology instead of utilizing conventional text mining features), it is not easily applicable to the general text classification problem. Here, we introduce a different approach using word embeddings and persistent homology which can be applied to the general text classification tasks.

\section{Methodology}
\label{sec:method}

In our algorithm, \emph{Topological Inference of the Embedding Space} (TIES) we utilize word embedding and persistent homology. The input is the textual document and the output is a topological representation of the same document. Later we may use these representations for text classification, clustering, etc. Step-by-step specifications of TIES are explained is this section.

\subsection{Pre-processing}

Like any other text mining method, standard pre-processing possibly including lemmatization, removing stop words  and if necessary lowercasing will be applied to the text. Also, there might be some specific pre-processing tasks that are inspired by TDA algorithms.

\subsection{Word Embedding Representation}

In a document of size $T$, replacing each token with its embedding vector of size $D$ will result in a matrix $X$ of size $T\times{D}$. This matrix can naturally be viewed as a $D$-dimensional time series that represents the document. More precisely, each column $X_d$ ($d=1,\dots,D$) will represent the document in dimension $d$ of the embedding.

\subsection{Aggregation on Sliding Window}

One of the easiest and potentially most efficient ways of such smoothing is to replace each element $X_{t,d}$ ($t=1,\dots,T$) in $X_d$ with a local average in its neighborhood. Equivalently, we may describe it by taking the summation in the sliding window of size $\omega$, where $c = (\omega-1)/2$ so the smoothed vector is
$$\tilde{X}_{t,d} = X_{t-c,d} + X_{t-c+1,d} + \dots + X_{t,d} + \dots +  X_{t+c,d}$$
and $\tilde{X}_{(T-\omega+1)\times{D}}$ is the smoothed $D$-dimensional time series that represents the document. For long documents this is almost the same as using a  smoother matrix $S_{T\times{T}}$, i.e.,
$$\tilde{X}_{T\times{D}} = S_{T\times{T}} X_{T\times{D}}$$
where $S$ is a tridiagonal (for $\omega=3$), pentadiagonal (for $\omega=5$) or heptadiagonal (for $\omega=7$) binary matrix. The only difference is that in the latter definition, no value of time series will be dropped, so the result is only slightly different in size, assuming $T >> \omega$. Note that we can also use exponential weights in summation of elements in the sliding window instead of simply adding them up. In one of our experiments, we tried the exponential form of:

\begin{equation}
    \notag
    \begin{split}
        \tilde{X}_{t,d} &= \frac{1}{8}X_{t-3,d} + \frac{1}{4}X_{t-2,d} + \frac{1}{2}X_{t-1,d} + X_{t,d}\\ &+ \frac{1}{2}X_{t+1,d} + \frac{1}{4}X_{t+2,d} + \frac{1}{8}X_{t+3,d}
\end{split}
\end{equation}

\subsection{Computing Distances}\label{step4}

Assume that in a documents, some of the embedding dimensions--- and the relation among different dimensions are carrying some information regarding the document. Such information could be revealed in a coordinate system, where each embedding dimension is represented by a data point, and the distance between two data points (embedding dimensions) represents their relation. A possible choice of distance is formulated in Equation \ref{eq:EQ_p1}.

\begin{equation}
    \begin{split}
    \varphi(\tilde{X}_i , \tilde{X}_j):&= \sqrt{\mathbb{E}[\tilde{X}_i^2]\mathbb{E}[\tilde{X}_j^2]} - \mathbb{E}[\tilde{X}_i \tilde{X}_j]\\
       &= \frac{1}{T} \norm{\tilde{X}_i}\norm{\tilde{X}_j} - \frac{1}{T}\tilde{X}_i^T \tilde{X}_j\\ 
       &= \frac{1}{T} \norm{\tilde{X}_i}\norm{\tilde{X}_j} \{ 1 - CosSim(\tilde{X}_i , \tilde{X}_j) \}
        \label{eq:EQ_p1}
    \end{split}
\end{equation}

The intuition behind the way of defining distance in Equation \ref{eq:EQ_p1} is to (1) consider the relation between $\tilde{X}_i$ and $\tilde{X}_j$ via Cosine similarity, (2) distinguish significant embedding dimensions (the term $\norm{\tilde{X}_i}\norm{\tilde{X}_j}$ will do this), and (3) make the distance almost non-sensitive to the size of document via term $1/T$. Note that Equation \ref{eq:EQ_p1} can be replaced by any other definition satisfying these three conditions. Aggregation on sliding window along with a distance formula like Equation \ref{eq:EQ_p1} guarantee that the order of the tokens in documents is considered in the final distance matrix $\Phi$, defined by
$$\Phi_{D\times{D}} = [\varphi(\tilde{X}_i , \tilde{X}_j)] \hspace{15pt};\hspace{15pt} i,j = 1,\dots,D.$$

For simplicity let's fix $\omega = 5$, so remembering that each column of $\tilde{X}$ as a simple time series is  a function of time (the index of word/token in the document),

$$\tilde{X}_{i} = \tilde{X}_i(t) = X_i(t-2) + X_i(t-1) + \dots + X_i(t+2)$$

and assuming that the length of document $T$ is large enough ($T >>1$) we have $$\mathbb{E}[X_i(t+s) X_j(t+s)] \rightarrow \mathbb{E}[X_i(t) X_j(t)]$$ as $T \rightarrow \infty$, so
$$\mathbb{E}[X_i(t+s) X_j(t+s)] \approx \mathbb{E}[X_i(t) X_j(t)]$$
since shifting the time index will only exclude $|s|$ elements from the beginning or the end of time series and its effect is negligible when $T >> |s|$.  It is easy to show that
\begin{equation}
    \notag
    \begin{split}
\mathbb{E}[\tilde{X}_i \tilde{X}_j] &=  \mathbb{E}[\tilde{X}_i(t) \tilde{X}_j(t)]\\
                                   &\approx 5 \mathbb{E} [X_i(t) X_j(t)]\\
                                   &+ 4 \mathbb{E} [X_i(t-1) X_j(t)] + 4 \mathbb{E} [X_i(t) X_j(t-1)]\\
                                   &+ 3 \mathbb{E} [X_i(t-2) X_j(t)] + 3 \mathbb{E} [X_i(t) X_j(t-2)]\\
                                   &+ 2 \mathbb{E} [X_i(t-3) X_j(t)] + 2 \mathbb{E} [X_i(t) X_j(t-3)]\\
                                   &+ 1 \mathbb{E} [X_i(t-4) X_j(t)] + 1 \mathbb{E} [X_i(t) X_j(t-4)]\\
    \end{split}
\end{equation}

and similarly, in a general form for any window size $\omega$, Equation \ref{eq:EQ_p2} holds.

\begin{equation}
\begin{split}
\mathbb{E} [\tilde{X}_i \tilde{X}_j] &=  \mathbb{E}[\tilde{X}_i(t) \tilde{X}_j(t)]\\
&\rightarrow w \mathbb{E} [X_i(t) X_j(t)]\\
&+ (\omega-1) \mathbb{E} [X_i(t-1) X_j(t)] + (\omega-1) \mathbb{E} [X_i(t) X_j(t-1)]\\
&+ (\omega-2) \mathbb{E} [X_i(t-2) X_j(t)] + (\omega-2) \mathbb{E} [X_i(t) X_j(t-2)]\\
& \hspace{25pt} \vdots  \hspace{45pt} \vdots \hspace{55pt} \vdots  \hspace{45pt} \vdots\\
&+ 1 \hspace{5pt} \mathbb{E} [X_i(t-\omega+1) X_j(t)] + 1 \hspace{5pt} \mathbb{E} [X_i(t) X_j(t-\omega+1)]\\
\text{as } &  \text{$T \rightarrow \infty$}
    \label{eq:EQ_p2}                                   
    \end{split}
\end{equation}

Such coefficients will guarantee that the order is considered in the final distance matrix $\Phi$. It means that each embedding dimension for each token in the text is being compared with all the other embedding dimensions in the same token, a few tokens before that, and a few tokens after that, though these comparisons will have different weights. Note that similar equations can be easily derived for correlation-based and covariance-based distances. In any case, the distance matrix is sensitive to the window size $\omega$, or more generally to the smoothing algorithm. For instance, using exponential smoothing will result in geometric sequence of coefficients instead of arithmetic sequence of coefficients in Equation \ref{eq:EQ_p2} (i.e., $\omega$, $\omega-1$, $\dots$,$1$). Regarding using the sliding window, the choice of $\omega$ is a trade-off between increasing the captured information on one side and decreasing the noise on the other side. 
   
\subsection{Applying Persistent Homology}

Having the distance matrix $\Phi$ for each document, a persistence diagram $PD(\Phi)$ can be constructed for topological dimension\footnote{These dimensions should not be mistaken with embedding dimensions.}  $0$ (number of clusters) and dimension $1$ (number of loops) denoted by $PD_0(\Phi)$ and $PD_1(\Phi)$ respectively. However, this persistence diagram alone is not very useful as the representation of the document.

In our prior work \cite{gholizadeh2018topological} the resulted graph of the main characters (persons) in each novel, and therefore the distance matrix was not annotated nor was needed to be annotated, since we had designed the algorithm to deal with the main characters whatever their names are. In other word, to capture the topological signature of a novelist, it did not matter whether the names of the main characters are shifted. But here, dealing with time-series in different dimensions, the order of embedding dimensions is meaningful, since different embedding dimensions have different roles in representing document. Therefore, feeding the time series to the persistent homology algorithm is meaningless, unless we somehow manage to distinguish different dimensions. One intuitive way is comparing the persistence diagram with and without each embedding dimension. In other word we can measure the change in persistence diagram when we exclude one embedding dimension. We measure the sensitivity of the persistence diagram generated by Ripser \cite{bauer2019ripser, bauer2017ripser} to each embedding dimension to use it later as a measure of the sensitivity of the document itself to each embedding dimension.
This way the document will be represented in an array of size array of $D$ based on $PD_0(\Phi)$ and another array of size $D$ based on $PD_1(\Phi)$, as formulated in Equation \ref{eq:EQ__p3}, where $\Psi$ is any measure of distance between two persistence diagram, e.g, Wasserstein distance.

% \begin{footnotesize}
\begin{equation}
\label{eq:EQ__p3} 
\begin{split}
\mathbb{V}_{0,d} &= \Psi(\hspace{3pt}PD_0(\Phi)\hspace{6pt}, \hspace{6pt} PD_0(\Phi \setminus d)\hspace{3pt}) \hspace{10pt}; \hspace{10pt}d= 1,\dots,D \\
\mathbb{V}_{1,d} &= \Psi(\hspace{3pt}PD_1(\Phi)\hspace{6pt}, \hspace{6pt} PD_1(\Phi \setminus d)\hspace{3pt}) \hspace{10pt}; \hspace{10pt}d= 1,\dots,D
\end{split}
\end{equation}       
% \end{footnotesize}

\begin{figure*}[!ht]
\centering
\includegraphics[width = 0.999\textwidth]{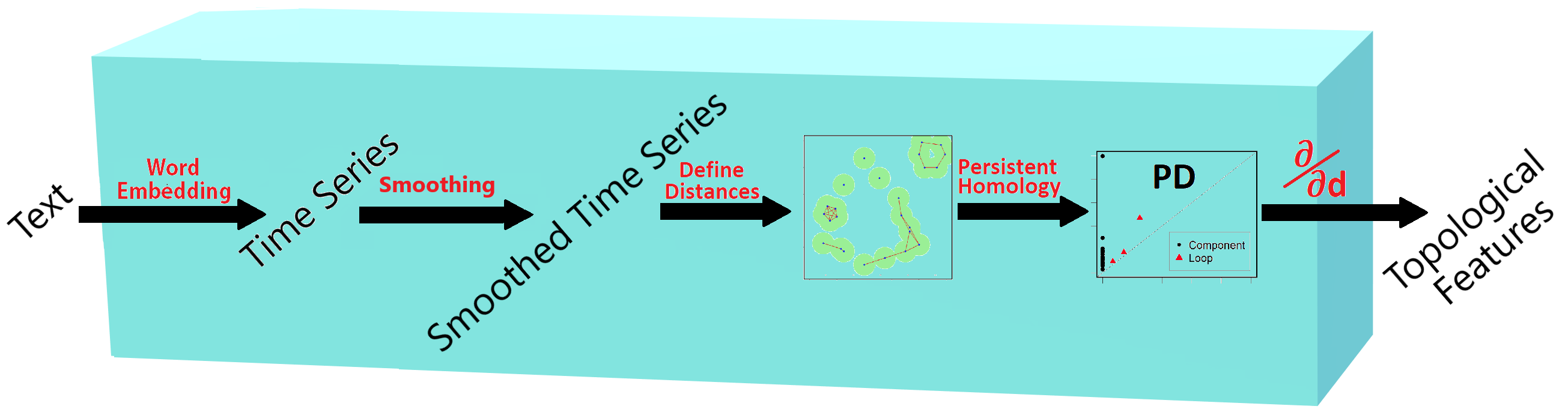}
\caption{A block diagram of TIES. Word Embedding representations are aggregated over sliding windows. Then defining the distance between pairs of embedding dimensions, persistence diagram can represent the text. Finally, the topological features are the sensitivity of the persistence diagram with respect to different dimensions.}
\label{fig:blockdiagram}
\end{figure*}

A block diagram of TIES is shown in Fig. \ref{fig:blockdiagram}.

\section{Data Specification}
\label{sec:data}

To examine our topological algorithm (TIES), we use the following data sets and predict the labels in multi-class multi labeling classification.

\begin{itemize}
    \item \textbf{arXiv Papers}: We downloaded all of arXiv papers in quantitative finance\footnote{https://arXiv.org/archive/q-fin} published between 2011 and 2018. Then we selected five major categories (subject tags): ``q-fin.GN'' (General Finance), ``q-fin.ST'' (Statistical Finance), ``q-fin.MF'' (Mathematical Finance), ``q-fin.PR'' (Pricing of Securities), and ``q-fin.RM'' (Risk Management). For pre-processing we removed the titles, author names and affiliations, abstracts, keywords and references. Then we tried to predict the subjects solely based on the paper body.\\
    
    \item \textbf{IMDB Movie Review} \cite{maas2011learning}: Using IMDB reviews annotated by positive/negative label, we examined TIES for binary sentiment classification task.  
\end{itemize}

\begin{figure}[!ht]
\centering
\includegraphics[width = 0.60\textwidth]{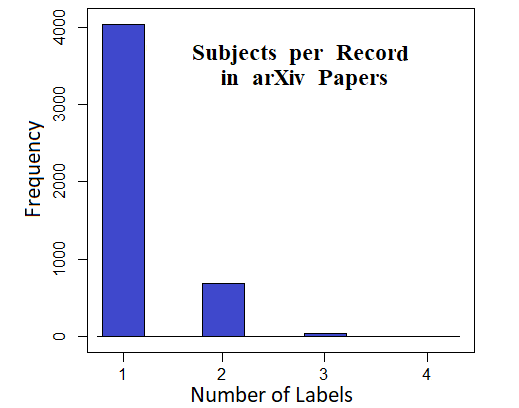}
\caption{Histograms of number of labels per document in arXiv data set of papers.}
\label{fig:hist}
\end{figure}

Table \ref{tab:data} contains the specifications of both data sets. Note that each records in the arXiv data set may have more that one label. The histogram of number of labels for each record is shown in Fig. \ref{fig:hist}. As shown in the histogram, the majority of records in arXiv data set are tagged with only a single label.

\begin{table}[ht]
    \caption{Data Specification for arXiv papers data sets.}
\label{tab:data}
    \centering
    \resizebox{.7\linewidth}{!}{
    \begin{tabular}{l c c}
    \hline
Specification & arXiv Quant. Fin. Papers & IMDB Movie Reviews \\ \hline
 Labels & 5 (Multi-label) &  2\\
 Clean Records & 4601 & 6000\\
 Length of Records  & $8456.9 \pm 6395.8$ & $540.5 \pm 171.9$ \\
 & & \\
 Frequency of Labels &  
 $\begin{matrix*}[l]
q\text{-}fin.GN: & 1258 \\
q\text{-}fin.ST: & 1144 \\
q\text{-}fin.MF: & 977 \\
q\text{-}fin.PR: & 907 \\
q\text{-}fin.RM: & 913 
\end{matrix*}$

&
$\begin{matrix*}[l]
Positive: & 3000 \\
Negative: & 3000 
\end{matrix*}$

\\ \hline
    \end{tabular}
    }
\end{table}

In practice, for many of the classification and clustering tasks in text processing, the data covers only very short documents (e.g., a limited data set of social media posts). Therefore a big challenge is training word embedding models on short documents. Such a challenge is beyond the scope of this study and we will use pre-trained versions of word embeddings that are previously trained on large corpora. Specifically, we use the following pre-trained models. 

\begin{itemize}
    \item GloVe \cite{pennington2014glove} pre-trained on Wikipedia 2014 and Gigaword 5 with vocabulary size of $400$K and $300$d vectors\footnote{http://nlp.stanford.edu/data/glove.6B.zip}.

    \item fastText \cite{bojanowski2016enriching, joulin2016bag} pre-trained on Wikipedia 2017 with the vocabulary size of $1$M and $300$d vectors\footnote{https://dl.fbaipublicfiles.com/fasttext/vectors-wiki/wiki.en.vec}.
    \item ConceptNet Numberbatch \cite{speer2017conceptnet} $v17.06$ with the vocabulary size of $400$K and $300$d vectors\footnote{https://conceptnet.s3.amazonaws.com/downloads/2017/numberbatch/numberbatch-en-17.06.txt.gz}.

\end{itemize}

\begin{table}[ht]
    \caption{Results on arXiv papers dataset. The best result is achieved using ConceptNet Numberbatch as pre-trained embedding and window size of 3.}
\label{tab:result_arXiv}
    \centering
    \begin{tabular}{c l c c c c c  }
    \hline
Model & Embedding  &  Window  &  Prec.  &  Rec. &  F1  &  Acc.\\ \hline

TIES + XGBoost & fastText  &  3  &  61.9  &  55.4  &  0.575  &  80.1\\
TIES + XGBoost & GloVe  &  3  &  63.1  &  56.7  &  0.597  &  80.7\\
TIES + XGBoost & Numberbatch  &  3  &  \textbf{68.7}  &  \textbf{60.5}  &  \textbf{0.643}  &  \textbf{82.6}\\
TIES + XGBoost & fastText  &  5  &  60.8  &  54.7  &  0.576  &  79.8\\
TIES + XGBoost & GloVe  &  5  &  61.8  &  56.1  &  0.588  &  80.3\\
TIES + XGBoost & Numberbatch  &  5  &  65.5  &  58.4  &  0.617  &  81.6\\
TIES + XGBoost & fastText  &  7  &  58.9  &  54.4  &  0.566  &  79.5\\
TIES + XGBoost & GloVe  &  7  &  62.8  &  56.4  &  0.594  &  80.6\\
TIES + XGBoost & Numberbatch  &  7  &  65.7  &  57.7  &  0.614  &  81.3\\
TIES + XGBoost & fastText  &  7 expon.  &  60.3  &  54.6  &  0.573  &  79.7\\
TIES + XGBoost & GloVe  &  7 expon.  &  61.2  &  55.9  &  0.584  &  80.2\\
TIES + XGBoost & Numberbatch  &  7 expon.  &  66.4  &  59.6  &  0.628  &  82.2\\ \hline
CNN & fastText & - & 57.1 & 64.3 & 60.5 & 80.0
\\
CNN & GloVe & - & 57.6 & 64.2 & 60.7 & 80.6
 \\
CNN & Numberbatch & - & \textbf{55.0} & \textbf{67.6} & \textbf{60.7} & \textbf{79.8}
 \\
\hline
    \end{tabular}
\end{table}

\begin{table}[ht]
    \caption{Results on IMDB Movie Review dataset. The best result is achieved using ConceptNet Numberbatch as pre-trained embedding and window size of 3.}
\label{tab:result_imdb}
    \centering
    \begin{tabular}{ c l c c c c c }
    \hline
Model & Embedding  &  Window  &  Prec.  &  Rec.  &  F1  &  Acc. \\ \hline
TIES + XGBoost & fastText     &  3  &  84.8  &  85.8  &  0.853  &  85.4\\
TIES + XGBoost & GloVe        &  3  &  86.9  &  88.0  &  0.874  &  87.5\\
TIES + XGBoost & Numberbatch  &  3  &  \textbf{87.9}  &  \textbf{89.0}  &  \textbf{0.884}  &  \textbf{88.5}\\
TIES + XGBoost & fastText     &  5  &  84.2  &  85.2  &  0.847  &  84.8\\
TIES + XGBoost & GloVe        &  5  &  85.6  &  86.6  &  0.861  &  86.2\\
TIES + XGBoost & Numberbatch  &  5  &  86.5  &  87.6  &  0.870  &  87.1\\
TIES + XGBoost & fastText     &  7  &  82.8  &  83.8  &  0.833  &  83.4\\
TIES + XGBoost & GloVe        &  7  &  83.8  &  84.8  &  0.843  &  84.4\\
TIES + XGBoost & Numberbatch  &  7  &  85.3  &  86.3  &  0.858  &  85.9\\
TIES + XGBoost & fastText     &  7 expon.  &  84.3  &  85.3  &  0.848  &  84.9\\
TIES + XGBoost & GloVe        &  7 expon.  &  86.5  &  87.6  &  0.870  &  87.1\\
TIES + XGBoost & Numberbatch  &  7 expon.  &  87.0  &  88.1  &  0.875  &  87.6 \\ \hline
Shauket et al. (2020) \cite{shaukat2020sentiment} & Lexicon based & - &   &   &   & 86.7 \\
Giatsoglou et al. (2017) \cite{giatsoglou2017sentiment} & Hybrid & - &   &   & \textbf{0.880} & \textbf{87.8} \\
\hline 
    \end{tabular}
\end{table}

\begin{table}[ht]
    \caption{Results per class on arXiv papers dataset using ConceptNet Numberbatch as pre-trained embedding and window size of 3.}
\label{tab:best_arXiv}
    \centering
    \begin{tabular}{ l c c c c c }
    \hline
Subject  &  Test Records  &  Precision  &  Recall  &  F1  &  Accuracy \\ \hline
q-fin.GN  &  410  &  73.2  &  68.5  &  0.708  &  83.8 \\
q-fin.ST  &  396  &  70.2  &  67.5  &  0.688  &  83.6 \\
q-fin.MF  &  306  &  66.0  &  45.6  &  0.539  &  77.5 \\
q-fin.PR  &  305  &  69.5  &  55.2  &  0.615  &  82.7 \\
q-fin.RM  &  307  &  62.5  &  61.0  &  0.617  &  84.5 \\ \hline

    \end{tabular}
\end{table}

\section{Results and Discussion}
\label{sec:results}
We run our binary classification and multi-label multi-class classification on both data set using XGBoost \cite{chen2015xgboost, chen2016xgboost} with the parameters $eta=0.25$, $max\_depth = 10$, $subsample = 0.5$, and $colsample\_bytree=0.5$. In each data set 2/3 of the records were randomly selected for training and 1/3 used for testing. Table \ref{tab:result_arXiv} and Table \ref{tab:result_imdb} show the results on arXiv paper data set and IMBD movie review data set, respectively. On each data set, we run the classifier using different pre-trained embedding models and different sliding window sizes to smooth the embedding signals. For both arXiv papers set and IMDB Movie Review data set, the best result is achieved using ConceptNet Numberbatch as pre-trained embedding and window size of $3$. Detailed results for arXiv papers set are shown in Table \ref{tab:best_arXiv}. 

To evaluate out results, for arXiv data set we run a convolutional neural network using the same pre-trained word embeddings. As shown in Table \ref{tab:result_arXiv}, our best configuration using TIES outperforms the base CNN model in terms of accuracy and F1 score.
For IMDB reviews data set, we compare our results to the previous results of Shauket et al. (2020) \cite{shaukat2020sentiment} lexicon based approach and Giatsoglou et al. (2017) \cite{giatsoglou2017sentiment} hybrid approach. The comparison reveals that TIES outperforms the previous models.

\section{Conclusion}
In this paper, we introduced a novel method to define and extract topological features from word embedding representations of corpus and used them for text classification. We utilized persistent homology, the most commonly tool from topological data analysis to interpret the embedding space of each textual documents. In our experiments, we showed that working on textual documents, our defined topological features can outperform conventional text mining features. Specially when the textual documents are long, using these topological features can improve the results. However, in TIES, we are analyzing different embedding dimensions as time series. Thus, to achieve reasonable results, a large number of tokens in each textual document is required. We acknowledge this issue as the main limitation of our algorithm. Also, it is not easy to measure and/or interpret the impact of different parts of the text input on the output in TIES. This is one of the possible future directions for this study.

\bibliographystyle{unsrt}
\bibliography{main}

\end{document}